\documentclass{article}%
\usepackage{amsmath}
\usepackage{amsfonts}
\usepackage{amssymb}
\usepackage{graphicx}
\usepackage{algorithmic}
\usepackage{algorithm}
\usepackage{hyperref}%
\providecommand{\U}[1]{\protect \rule{.1in}{.1in}}
\providecommand{\U}[1]{\protect \rule{.1in}{.1in}}

\begin{document}

\title{Ensemble-Based Survival Models with the Self-Attended Beran Estimator Predictions}
\author{Lev V. Utkin, Semen P. Khomets,Vlada A. Efremenko,
\and Andrei V. Konstantinov and Natalya M. Verbova\\Higher School of Artificial Intelligence Technologies\\Peter the Great St.Petersburg Polytechnic University\\St.Petersburg, Russia\\e-mail: utkin\_lv@spbstu.ru, homets\_sp@spbstu.ru,\\efremenko\_va@spbstu.ru, konstantinov\_av@spbstu.ru,\\verbova\_nm@spbstu.ru}
\date{}
\maketitle

\begin{abstract}
Survival analysis predicts the time until an event of interest, such as
failure or death, but faces challenges due to censored data, where some events
remain unobserved. Ensemble-based models, like random survival forests and
gradient boosting, are widely used but can produce unstable predictions due to
variations in bootstrap samples. To address this, we propose SurvBESA
(Survival Beran Estimators Self-Attended), a novel ensemble model that
combines Beran estimators with a self-attention mechanism. Unlike traditional
methods, SurvBESA applies self-attention to predicted survival functions,
smoothing out noise by adjusting each survival function based on its
similarity to neighboring survival functions. We also explore a special case
using Huber's contamination model to define attention weights, simplifying
training to a quadratic or linear optimization problem. Numerical experiments
show that SurvBESA outperforms state-of-the-art models. The implementation of
SurvBESA is publicly available.

\end{abstract}

\section{Introduction}

Survival analysis is a vital field focused on predicting the time until an
event of interest occurs \cite{Wang-Li-Reddy-2019}. A key challenge in this
domain is the presence of censored data, where some events remain unobserved
during the study period, resulting in observed times that are less than or
equal to the true event times \cite{Nezhad-etal-2018}. As a result, survival
datasets typically contain a mix of censored and uncensored observations,
requiring specialized methods to handle such data effectively.

One of the most powerful approaches for survival analysis is the use of
ensemble-based models. These models combine multiple weak or base models to
create a robust, accurate, and generalizable predictive model. Ensemble
methods have been extensively applied and refined in survival analysis, often
as extensions of well-established techniques. For instance, Random Survival
Forests (RSFs)
\cite{Ibrahim-etal-2008,Mogensen-etal-2012,Schmid-etal-2016,Wang-Zhou-2017,Utkin-etal-2019c,Utkin-Konstantinov-22d,Wright-etal-2017}
adapt Random Forests to survival data, while gradient boosting-based survival
models \cite{chen2013gradient,liu2019hitboost} extend Gradient Boosting
Machines. The key differences among these models lie in the choice of weak
learners and the strategies used to aggregate their predictions. An innovative
approach by Meier et al. \cite{Meier-etal-16} employs the Cox proportional
hazards model \cite{Cox-1972} as a weak learner. While the Cox model is a
cornerstone of survival analysis, it has limitations: it estimates conditional
survival measures using feature vectors but ignores their relative positions,
leading to inaccuracies in datasets with multi-cluster structures. To address
this, the Beran estimator \cite{Beran-81} is used. This estimator calculates
the conditional survival function by weighting instances based on their
proximity to the analyzed instance using kernel functions, effectively
performing kernel regression.

Extensive surveys and discussions on ensemble-based approaches and their
applications can be found in
\cite{Xibin_Dong-etal-20,Ferreira-Figueiredo-2012,Ren-Zhang-Suganthan-2016,Sagi-Rokach-2018,Wozniak-etal-2014,ZH-Zhou-2012}%
.

The Beran estimator's ability to incorporate relationships between feature
vectors makes it an attractive choice as a weak learner in new ensemble-based
models. A simple implementation involves averaging predictions from multiple
Beran estimators using bagging. However, predictions from different weak
models, in the form of survival functions (SFs), can vary significantly,
particularly when trained on data from different bootstrap samples. This may
lead to incorrect and unstable aggregation of the base model predictions,
resulting in an unreliable aggregated SF for the ensemble.

We propose to aggregate the survival function predictions by applying the
self-attention mechanism \cite{Vaswani-etal-17}, which allows a model to weigh
the importance of different elements in a sequence relative to each other. The
main idea behind using self-attention is to capture dependencies among the
predictions provided by the different Beran models in the ensemble and to
remove noise or anomalous predictions that may arise due to a potentially
\textquotedblleft unfortunate\textquotedblright \ selection of a subset of
training examples during the bootstrap sampling process used to construct a
base Beran estimator. In contrast to the self-attention mechanism in
transformer-based survival models
\cite{hu2021transformer,tang2023explainable,Wang-Sun-22}, which use
self-attention to weigh feature vectors from a training set, the proposed
model aims to weigh the SFs predicted by the base Beran estimators from the
ensemble. In other words, we propose to adjust the predicted SFs using
self-attention based on the neighboring SFs. Specifically, the feature vectors
in the conventional self-attention framework are replaced with predicted SFs.
As a result, anomalous or noisy SFs are smoothed in accordance with the
nearest SFs. The proposed model is called SurvBESA (Survival Beran Estimators Self-Attended).

We also consider an interesting special case where the self-attention weights
are defined using Huber's $\epsilon$-contamination model \cite{Huber81} with
parameter $\epsilon$ and its imprecise extension \cite{Walley91}. A key
feature of this case is that the training task is reduced to a quadratic or
linear optimization problem. Similar approaches applying the $\epsilon
$-contamination model have been used in
\cite{Utkin-Konstantinov-22,Utkin-Konstantinov-Kirpichenko-23}.

Numerical experiments comparing SurvBESA with other survival models, including
Random Survival Forests (RSF) \cite{Ibrahim-etal-2008}, GBM Cox
\cite{ridgeway1999state}, GBM AFT \cite{Barnwal-etal-22}, and the standalone
Beran estimator.

The implementation of SurvBESA is publicly available at: \url{https://github.com/NTAILab/SurvBESA}.

The paper is organized as follows. Related work reviewing papers devoted to
machine learning in survival analysis and to the self-attention mechanism can
be found in Section 2. A short description of the survival analysis concepts
is given in Section 3. An idea of SurvBESA as an attention-based ensemble of
Beran estimators is provided in Section 4. Numerical experiments comparing
different survival models trained on synthetic and real data are considered in
Section 5. Concluding remarks can be found in Section 6.

\section{Related work}

\textbf{Machine learning models in survival analysis}. Comprehensive overviews
of survival models can be found in \cite{Wang-Li-Reddy-2019, Wiegrebe:2024aa}.
Many of these models are extensions of standard machine learning techniques
adapted to handle survival data. In recent years, neural networks and deep
learning have gained traction in survival analysis. Notable contributions
include deep survival models \cite{Katzman-etal-2018, Luck-etal-2017,
Nezhad-etal-2018, Yao-Zhu-Zhu-Huang-2017}, a deep recurrent survival model
\cite{ren2019deep}, convolutional neural networks for survival tasks
\cite{Haarburger-etal-2018}, and transformer-based survival models
\cite{hu2021transformer, tang2023explainable, Wang-Sun-22}.

However, survival models based on neural networks often require large amounts
of training data, which can be a significant limitation in many real-world
applications. Therefore, ensemble-based survival models that utilize simpler
weak learners may offer more robust and accurate predictions, particularly
when data is limited or high-dimensional. Basic principles of survival
ensembles was presented by Hothorn et al. \cite{Hothorn-etal-2006}. Various
ensemble-based survival models were also presented in
\cite{chen2013gradient,liu2019hitboost,Meier-etal-16,Mogensen-etal-2012,Schmid-etal-2016,Wang-Zhou-2017,Utkin-etal-2019c,Utkin-Konstantinov-22d,Wright-etal-2017}%
. However, it is interesting to point out that there are no ensemble-based
models which use the Beran estimator as a weak model. At the same time, our
experiments show that models based on the Beran estimators can compete with
other models in case of complex data structure.

\textbf{Self-attention}. The self-attention mechanism, introduced by Vaswani
et al. \cite{Vaswani-etal-17}, is a key component of the Transformer neural
network architecture. It builds on earlier works, such as Cheng et al.
\cite{ChengDong-Lapata-16} and Parikh et al. \cite{Parikh-etal-16}. Since its
introduction, it has been widely adopted across various domains, including
sentence embedding \cite{Lin-Feng-etal-17}, machine translation and natural
language processing \cite{Devlin-etal-18,Wu-Fan-etal-19}, speech recognition
\cite{Povey-etal-18,Shim-Choi-Sung-22}, and image recognition
\cite{Chen-Xie-etal-20,Dosovitski-etal-20,Guo-Liu-etal-21,Liu-Lin-etal-21,Shen-Bello-etal-20,soydaner2022attention,Wang-Jiang-etal-17,Wang-Girshick-etal-18,Zhao-Jia-Koltun-20}%
.

Numerous survey papers have explored the diverse aspects and applications of
attention and self-attention mechanisms, including
\cite{brauwers2021general,Lin-Wang-etal-21,Goncalves-etal-2022,hassanin2024visual,Khan-Naseer-etal-22,Santana-Colombini-21,Soydaner-22,Xu-Wei-etal-22}%
.

The self-attention mechanism implemented in a deep gated neural network for
survival analysis was introduced in \cite{yang2024deep}. The idea behind using
the self-attention mechanism in the proposed neural network is that it treats
time as an additional input covariate, with the condition that the smaller the
time interval, the higher the attention weight. SurvBESA, however, uses
self-attention differently. It serves as an aggregation operation for the
predictions of the Beran estimators.

\subsection{Basic concepts of survival analysis}

The dataset $\mathcal{A}$ is assumed to consist of $n$ vectors of the form
$(\mathbf{x}_{i},\delta_{i},T_{i})$, where $i=1,...,n$. Here $\mathbf{x}%
_{i}^{\mathrm{T}}\in \mathbb{R}^{d}$ represents the feature vector for the
$i$-th object, $T_{i}\in \mathbb{R}_{+}$ is the event time, and $\delta_{i}%
\in \{0,1\}$ is the censoring indicator. Specifically, $\delta_{i}=1$ indicates
that the event is observed (uncensored), while $\delta_{i}=0$ corresponds to a
censored observation \cite{Hosmer-Lemeshow-May-2008}. Given the dataset
$\mathcal{A}$, the goal is to construct a survival model capable of estimating
the event time $T$ for a new object $\mathbf{x}$.

This estimation can be represented by the conditional SF, denoted
$S(t\mid \mathbf{x})$, which is the probability of surviving beyond time $t$,
i.e., $S(t\mid \mathbf{x})=\Pr \{T>t\mid \mathbf{x}\}$. Alternatively, the
estimation can be expressed through the cumulative hazard function (CHF),
denoted as $H(t\mid \mathbf{x})$, which is related to the SF as $H(t\mid
\mathbf{x})=-\ln S(t\mid \mathbf{x})$. While other representations exist
\cite{Wang-Li-Reddy-2019}, we focus on these two.

A key question in survival analysis is how to compare the performance of
different survival models. One widely used metric is the C-index, introduced
by Harrell et al. \cite{Harrell-etal-1982}. The C-index estimates the
probability that the predicted event times of a pair of objects are correctly
ranked \cite{Wang-Li-Reddy-2019}. Let $\mathcal{J}$ denote the set of all
pairs $(i,j)$ of objects satisfying $\delta_{i}=1$ and $T_{i}<T_{j}$. The
C-index can then be computed as \cite{Uno-etal-11,Wang-Li-Reddy-2019}:%
\begin{equation}
C=\frac{1}{\# \mathcal{J}}\sum_{(i,j)\in \mathcal{J}}\mathbf{1}[\widehat{T}%
_{i}<\widehat{T}_{j}],
\end{equation}
where $\widehat{T}_{i}$ and $\widehat{T}_{j}$ are predicted event times for
objects with indices $i$ and $j$, respectively.

Let $t_{1}<t_{2}<...<t_{n}$ be an ordered sequence of times $\{T_{1}%
,...,T_{n}\}$. Then the SF can be estimated using the Beran estimator
\cite{Beran-81} as follows:
\begin{equation}
S(t\mid \mathbf{x},\mathcal{A})=\prod_{t_{i}\leq t}\left \{  1-\frac
{\alpha(\mathbf{x},\mathbf{x}_{i})}{1-\sum_{j=1}^{i-1}\alpha(\mathbf{x}%
,\mathbf{x}_{j})}\right \}  ^{\delta_{i}}, \label{Beran_est}%
\end{equation}
where the weight $\alpha(\mathbf{x},\mathbf{x}_{i})$ reflects the relevance of
the $i$-th object $\mathbf{x}_{i}$ to the feature vector $\mathbf{x}$. This
weight can be defined using a kernel function:
\begin{equation}
\alpha(\mathbf{x},\mathbf{x}_{i})=\frac{K(\mathbf{x},\mathbf{x}_{i})}%
{\sum_{j=1}^{n}K(\mathbf{x},\mathbf{x}_{j})}.
\end{equation}

For example, when the Gaussian kernel is used, the weights correspond to the
softmax operation with a temperature parameter $\tau$:
\begin{equation}
\alpha(\mathbf{x},\mathbf{x}_{i})=\text{\textrm{softmax}}\left(
-\frac{\left \Vert \mathbf{x}-\mathbf{x}_{i}\right \Vert ^{2}}{\tau}\right)  .
\end{equation}

Notably, the Beran estimator generalizes the Kaplan-Meier estimator. When all
weights $\alpha(\mathbf{x},\mathbf{x}_{i})$ are set to $1/n$ for $i=1,...,n$,
the Beran estimator reduces to the Kaplan-Meier estimator.

\section{Attention-based ensemble of Beran estimators}

\subsection{Ensemble of Beran estimators}

An ensemble of Beran estimators is constructed using the standard bagging
approach. This involves randomly selecting a subset $\mathcal{A}_{k}$ of $m$
objects from the dataset $\mathcal{A}$. Let $t_{1}^{(k)}<...<t_{m}^{(k)}$ be
the ordered event times of the objects in $\mathcal{A}_{k}$. The SF predicted
by the $k$-th Beran estimator trained on the subset $\mathcal{A}_{k}$ and
denoted as $S^{(k)}(t\mid \mathbf{x})$ is computed as follows:%
\begin{equation}
S(t\mid \mathbf{x},\mathcal{A}_{k})=S^{(k)}(t\mid \mathbf{x})=\prod_{t_{i}%
^{(k)}\leq t}\left \{  \frac{1-\sum_{j=1}^{i}\alpha(\mathbf{x},\mathbf{x}_{j}%
)}{1-\sum_{j=1}^{i-1}\alpha(\mathbf{x},\mathbf{x}_{j})}\right \}  ^{\delta_{i}%
}.
\end{equation}

SFs $S(t\mid \mathbf{x},\mathcal{A}_{k})$, where $k=1,...,M$, obtained using
$M$ Beran estimators are aggregated to produce the final SF as follows:
\begin{equation}
S(t\mid \mathbf{x})=\frac{1}{M}\sum_{k=1}^{M}S^{(k)}(t\mid \mathbf{x}).
\end{equation}

This model operates similarly to the standard random forest, with a key
distinction: the decision trees are replaced with Beran estimators. While
random forests aggregate predictions from multiple decision trees, this model
aggregates predictions from multiple Beran estimators, leveraging their
ability to handle survival data effectively.

\subsection{SurvBESA}

One of the challenges in using the simple averaging operation to aggregate
predictions from the Beran estimators is the potential for significant
variability in the predictions. Specifically, some weak learners may produce
anomalous survival functions (SFs) due to various factors, such as the
specific subset $\mathcal{A}_{k}$ used for training or the multi-cluster
structure of the data. These anomalous SFs can introduce bias into the final
predictions, leading to unreliable estimates. To address this issue, we
propose an alternative model designed to mitigate these problems and improve
the robustness of the predictions.

The first key idea behind the proposed model is to define attention weights
and new base SFs using the self-attention mechanism. This approach allows the
model to dynamically weigh the contributions of different base SFs based on
their relevance and relationships, improving the robustness and accuracy of
the final predictions.
\begin{equation}
\widetilde{S}^{(j)}(t\mid \mathbf{x},\mathbf{\theta})=\sum_{k=1,k\neq j}%
^{M}\beta \left(  S^{(j)}(t\mid \mathbf{x}),S^{(k)}(t\mid \mathbf{x}%
),\mathbf{\theta}\right)  \cdot S^{(k)}(t\mid \mathbf{x}), \label{Ens_Ber_18}%
\end{equation}
where $\beta(\cdot,\cdot,\cdot)$ is the self-attention weight which
establishes the relationship between SFs $S^{(j)}(t\mid \mathbf{x})$ and
$S^{(k)}(t\mid \mathbf{x})$; $\mathbf{\theta}$ is the vector of training or
tuning parameters.

In the context of the attention mechanism, $S^{(j)}(t\mid \mathbf{x})$ serves
as the query, while $S^{(k)}(t\mid \mathbf{x})$ functions as both the key and
the value. This setup allows the model to compute attention weights based on
the relationships between the predicted survival functions, enabling more
informed aggregation of the base predictions.

From the above, it follows that each base SF is adjusted using the
self-attention mechanism, taking into account the neighboring SFs. By applying
self-attention, we effectively reduce noise that may arise due to differences
in the subsets $\mathcal{A}_{k}$. To illustrate this, consider a scenario
where the data consists of two distinct clusters. Suppose a subset
$\mathcal{A}_{k}$ is randomly sampled from one cluster, but the instance
$\mathbf{x}$ being analyzed belongs to the other cluster. In this case, the
Beran estimator trained on $\mathcal{A}_{k}$ may incorrectly associate
$\mathbf{x}$ with the first cluster, as it lacks information about the second
cluster. This can lead to an inaccurate prediction of the SF $S^{(i)}%
(t\mid \mathbf{x})$. The self-attention mechanism addresses this issue by
incorporating contextual information from neighboring SFs. Specifically, each
new SF $\widetilde{S}^{(i)}(t\mid \mathbf{x})$ is computed as a weighted sum of
all SFs predicted by the $M-1$ other Beran estimators. The weights are
determined by the distance between the SFs, ensuring that closer (more
similar) SFs contribute more significantly to the final prediction. As a
result, the adjusted SF $\widetilde{S}^{(i)}(t\mid \mathbf{x})$ can be viewed
as a denoised and filtered version of the original prediction. This denoising
property of self-attention has been highlighted in \cite{Vidal-22} as
analogous to non-local means denoising, a technique designed to remove noise
from data by leveraging contextual information.

After obtaining the denoised SFs $\widetilde{S}^{(i)}(t\mid \mathbf{x}%
,\mathbf{\theta})$, $i=1,...,M$, we can compute the aggregated SF of the
ensemble as follows:%
\begin{equation}
\widetilde{S}(t\mid \mathbf{x},\mathbf{\theta})=\frac{1}{M}\sum_{k=1}%
^{M}\widetilde{S}^{(k)}(t\mid \mathbf{x},\mathbf{\theta}). \label{Ens_Ber_20}%
\end{equation}

The next question is how to define the attention weights $\beta$. To address
this, we first need to define a distance metric between two survival functions
(SFs). A well-known distance metric for probability distributions is the
Kolmogorov-Smirnov distance, which can be applied to a pair of SFs
$S^{(k)}(t\mid \mathbf{x})$ and $S^{(l)}(t\mid \mathbf{x})$ as:
\begin{equation}
D_{KS}(S^{(k)}(t\mid \mathbf{x}),S^{(l)}(t\mid \mathbf{x}))=\sup_{t}~\left \vert
S^{(k)}(t\mid \mathbf{x})-S^{(l)}(t\mid \mathbf{x})\right \vert .
\end{equation}

This distance measures the maximum absolute difference between the two SFs
over time $t$, providing a way to quantify their dissimilarity.

By using the Gaussian kernel, the attention weight $\beta$ is of the form:
\begin{align}
\beta(S^{(l)}(t  &  \mid \mathbf{x}),S^{(k)}(t\mid \mathbf{x}),\mathbf{\theta
})\nonumber \\
&  =\text{\textrm{softmax}}\left(  -\frac{D_{KS}(S^{(l)}(t\mid \mathbf{x}%
),S^{(k)}(t\mid \mathbf{x}))}{\theta_{kl}}\right)  ,
\end{align}
where $\theta_{kl}$ is a hyperparameter or training parameter.

We have $M(M-1)$ parameters $\theta_{kl}$, $k=1,...,M,$ $l=1,...,M$, $k\neq
l$. Hence, there holds $\mathbf{\theta}=(\theta_{12},\theta_{13}%
,...,\theta_{(M-1)M})$.

The simplest way for applying the self-attention and computing the aggregated
SF is to tune parameters $\tau$ and $\theta$ of kernels and to find the best
ones which provide the largest C-index. However, we can extend the set of
training parameters and maximize the C-index over these parameters.

By using the introduced notation $\mathcal{J}$ for the set of pairs $(i,j)$,
satisfying conditions $\delta_{i}=1$ and $T_{i}<T_{j}$, we write the C-index
as%
\begin{align}
C  &  =\frac{1}{N}\sum_{(i,j)\in \mathcal{J}}\mathbf{1}[\widehat{T}%
_{j}-\widehat{T}_{i}>0]\nonumber \\
&  =\frac{1}{N}\sum_{(i,j)\in \mathcal{J}}\mathbf{1}\left[  \sum_{k=1}%
^{M}\widehat{T}_{j}^{(k)}-\sum_{k=1}^{M}\widehat{T}_{i}^{(k)}>0\right]  .
\end{align}

Here $\widehat{T}_{i}^{(k)}$ is the expected time predicted by the $k$-th
Beran estimator for the $i$-th object. Since there is a finite number of
objects in the dataset, then the SF is step-wised. Hence, the conditional SF
predicted by the $k$-th Beran estimator is
\begin{equation}
S^{(k)}(t\mid \mathbf{x}_{i})=\sum \limits_{l=0}^{N_{k}-1}S_{l}^{(k)}%
(\mathbf{x}_{i})\cdot \mathbf{1}\{t\in \lbrack t_{l}^{(k)},t_{l+1}^{(k)})\},
\end{equation}
where $S_{l}^{(k)}(\mathbf{x}_{i})=S^{(k)}(t_{l}\mid \mathbf{x}_{i})$ is the SF
in the time interval $[t_{l}^{(k)},t_{l+1}^{(k)})$; $S_{0}^{(k)}=1$
by\ $t_{0}=0$; $N_{k}$ is the number of elements in $\mathcal{A}_{k}$;
$t_{1}^{(k)}<t_{2}^{(k)}<...t_{N_{k}}^{(k)}$ are ordered times to events from
$\mathcal{A}_{k}$.

Hence, $\widehat{T}_{i}^{(k)}$ is determined as
\begin{equation}
\widehat{T}_{i}^{(k)}=\sum \limits_{l=0}^{N_{k}-1}S_{l}^{(k)}(\mathbf{x}%
_{i})(t_{l+1}^{(k)}-t_{l}^{(k)}).
\end{equation}

Since $t_{1}^{(k)},t_{2}^{(k)},...,t_{N_{k}}^{(k)}$ is a part of all event
times, then we can use all times $t_{l}$, $l=1,...,N$, but if a time $t_{l}%
\in \lbrack t_{l-1},t_{l+1}]$ does not belong to the $k$-th bootstrap subset of
the event times, then $S_{l}^{(k)}(\mathbf{x}_{i})=S_{l-1}^{(k)}%
(\mathbf{x}_{i})$. It follows from (\ref{Ens_Ber_18}) and from the above
definition of $\widehat{T}_{i}^{(k)}$ that there holds%
\begin{equation}
\widehat{T}_{i}^{(j)}=\sum_{k=1,k\neq j}^{M}\beta \left(  S^{(j)}%
(t\mid \mathbf{x}_{i}),S^{(k)}(t\mid \mathbf{x}_{i}),\mathbf{\theta}\right)
\cdot \widehat{T}_{i}^{(k)}.
\end{equation}

Hence, the C-index is determined as%

\begin{align}
C  &  =\frac{1}{N}\sum_{(i,j)\in \mathcal{J}}\mathbf{1}[\widehat{T}%
_{j}-\widehat{T}_{i}>0]\nonumber \\
&  =\frac{1}{N}\sum_{(i,j)\in \mathcal{J}}\mathbf{1}\left[
\begin{array}
[c]{c}%
\sum_{l=1}^{M}\sum_{k=1,k\neq l}^{M}\beta \left(  S^{(l)}(t\mid \mathbf{x}%
_{j}),S^{(k)}(t\mid \mathbf{x}_{j}),\mathbf{\theta}\right)  \cdot \widehat
{T}_{j}^{(l)}\\
-\sum_{l=1}^{M}\sum_{k=1,k\neq l}^{M}\beta \left(  S^{(l)}(t\mid \mathbf{x}%
_{i}),S^{(k)}(t\mid \mathbf{x}_{i}),\mathbf{\theta}\right)  \cdot \widehat
{T}_{i}^{(l)}>0
\end{array}
\right]  .
\end{align}

Let us denote
\begin{equation}
\beta_{j}^{(l,k)}(\mathbf{\theta})=\beta \left(  S^{(l)}(t\mid \mathbf{x}%
_{j}),S^{(k)}(t\mid \mathbf{x}_{j}),\mathbf{\theta}\right)  ,
\end{equation}%
\begin{equation}
R_{ij}(\mathbf{\theta})=\sum_{l=1}^{M}\sum_{k=1,k\neq l}^{M}\left(  \beta
_{j}^{(l,k)}(\mathbf{\theta})\widehat{T}_{j}^{(k)}-\beta_{i}^{(l,k)}%
(\mathbf{\theta})\widehat{T}_{i}^{(k)}\right)  . \label{C-ind-func1}%
\end{equation}

Then there holds%
\begin{equation}
C=\frac{1}{N}\sum_{(i,j)\in \mathcal{J}}\mathbf{1}\left[  R_{ij}(\mathbf{\theta
})>0\right]  . \label{C-index-1}%
\end{equation}

It is proposed to replace the indicator function in the above C-index with the
sigmoid function $\sigma$. Hence, optimal parameters $\mathbf{\theta}$ of the
self-attention are obtained by solving the following optimization problem:%
\begin{equation}
\max_{\mathbf{\theta}}\sum_{(i,j)\in \mathcal{J}}\sigma \left(  R_{ij}%
(\mathbf{\theta})\right)  .
\end{equation}

Parameters $\mathbf{\theta}$ as well as the parameter of the Beran estimator
$\tau$ can be obtained by solving the above optimization problem by means of
gradient-based algorithms. Moreover, the parameter $\tau$ can be trained for
each base learner, i.e., we can have $M$ parameters $\tau_{1},...,\tau_{M}$
such that each parameter defines the corresponding Beran estimator. Finally,
the aggregation (\ref{Ens_Ber_20}) can also be replaced with the
Nadaraya-Watson kernel regression. However, this replacement may significantly
complicate the optimization problem, therefore, it is not considered in this work.

\subsection{An important special case}

If we suppose that $\eta$ is the hyperparameter which is identical for all
base learners, then the optimization problem for computing parameters
$\mathbf{\theta}$, can significantly be simplified. Let us use the definition
of the attention weight $\beta_{i}^{(k,j)}(\mathbf{\theta})$ in a form
proposed in \cite{Utkin-Konstantinov-22}:%
\begin{align}
\beta_{j}^{(l,k)}(\mathbf{\theta})  &  =(1-\epsilon)\cdot
\text{\textrm{softmax}}\left(  D_{j}^{(k,l)}(\varphi)\right)  +\epsilon
\cdot \theta_{l,k},\nonumber \\
k  &  =1,...,M,
\end{align}
where $\varphi$ is the tuning parameter; $\theta_{1,1},...,\theta_{M,M}$ are
training parameters such that $\mathbf{\theta}\in \Delta^{M\times M}$;
\[
D_{j}^{(k,l)}(\varphi)=-\frac{D_{KS}(S^{(k)}(T_{j}\mid \mathbf{x}_{j}%
),S^{(l)}(T_{j}\mid \mathbf{x}_{j}))}{\varphi}.
\]

The above expression is derived from the imprecise Huber's $\epsilon
$-contamination model \cite{Huber81}, which is represented as
\begin{equation}
F=(1-\epsilon)\cdot P+\epsilon \cdot \Theta,
\end{equation}
where $P$ is a discrete probability distribution contaminated by another
probability distribution $\Theta$, which is arbitrary within the unit simplex
$\Delta^{M\times M\prime}$; components of $\Theta$ satisfy the conditions
$\theta_{1,1}+...+\theta_{M,M}=1$ and $\theta_{l,k}\geq0$ for all $k=1,...,M$,
$l=1,...,M$; the contamination parameter $\epsilon \in \lbrack0,1]$ controls the
size of the small simplex generated by the $\epsilon$-contamination model. In
particular, if $\epsilon=0$, then the model reduces to the softmax operation
with hyperparameter $\varphi$, and the attention weight becomes independent of
$\theta_{l,k}$. The distribution $P$ consists of elements \textrm{softmax}%
$\left(  D_{j}^{(k,l)}(\varphi)\right)  $, $k=1,...,M$,\ $l=1,...,M$. The
distribution $\Theta$ defines the vector $\mathbf{\theta}=(\theta
_{1,1},...,\theta_{M,M})$, which represents the training parameters of the
attention mechanism.

Let us replace the indicator function in (\ref{C-index-1}) with the hinge loss
function $\max \left(  0,x\right)  $ similarly to the replacement proposed by
Van Belle et al. \cite{Van_Belle-etal-2007}. By adding the regularization term
$\left \Vert \mathbf{\theta}\right \Vert ^{2}$ $=\sum_{l=1,}^{M}\sum_{k=1,k\neq
l}^{M}\theta_{l,k}^{2}$ with the hyperparameter $\lambda$ which controls the
strength of the regularization, the optimization problem can be written as%
\begin{equation}
\min_{\mathbf{\theta}_{1}\in \Delta^{M},...,\mathbf{\theta}_{M}\in \Delta^{M}%
}\left \{  \sum_{(i,j)\in \mathcal{J}}\max \left(  0,R_{ij}(\mathbf{\theta
})\right)  +\lambda \left \Vert \mathbf{\theta}\right \Vert ^{2}\right \}  .
\end{equation}

Here $\mathbf{\theta}_{j}$ is the vector of $M$ variables $\theta
_{j,1},...,\theta_{j,M}$. Let us introduce the variables%
\begin{equation}
\xi_{ij}=\max \left(  0,R_{ij}(\mathbf{\theta})\right)  .
\label{Survival_DF_48}%
\end{equation}

The optimization problem can be written in the following form:
\begin{equation}
\min_{\mathbf{\theta}_{1}\in \Delta^{M},...,\mathbf{\theta}_{M}\in \Delta^{M}%
}\left \{  \sum_{(i,j)\in \mathcal{J}}\xi_{ij}+\lambda \left \Vert \mathbf{\theta
}\right \Vert ^{2}\right \}  , \label{Survival_DF_50}%
\end{equation}
subject to $\mathbf{\theta}\in \Delta^{M}$ and
\begin{equation}
\xi_{ij}\geq R_{ij}(\mathbf{\theta}),\  \  \xi_{ij}\geq0,\  \  \{i,j\}
\in \mathcal{J}. \label{Survival_DF_51}%
\end{equation}

Let us consider how $R_{ij}(\mathbf{\theta})$ depends on $\mathbf{\theta}$ in
this special case. Denote%
\begin{gather}
Q_{i,j}^{(k,l)}(\varphi,\epsilon)=(1-\epsilon)\cdot S^{(k)}(T_{j}%
\mid \mathbf{x}_{j})\cdot \text{\textrm{softmax}}\left(  D_{j}^{(k,l)}%
(\varphi)\right) \nonumber \\
-(1-\epsilon)\cdot S^{(k)}(T_{i}\mid \mathbf{x}_{i})\cdot \text{\textrm{softmax}%
}\left(  D_{i}^{(k,l)}(\varphi)\right)  ,
\end{gather}
and
\begin{equation}
G_{i,j}^{(k)}(\epsilon)=\epsilon \cdot \left(  S^{(k)}(T_{j}\mid \mathbf{x}%
_{j})-S^{(k)}(T_{i}\mid \mathbf{x}_{i})\right)  .
\end{equation}

It follows from (\ref{C-ind-func1}) that
\begin{equation}
R_{ij}(\mathbf{\theta})=\sum_{l=1,}^{M}\sum_{k=1,k\neq l}^{M}\left(
Q_{i,j}^{(k,l)}(\varphi,\epsilon)+\theta_{l,k}\cdot G_{i,j}^{(k)}%
(\epsilon)\right)  . \label{C-ind-func2}%
\end{equation}

It can be seen from (\ref{C-ind-func2}) that the constraints are linear with
respect to training parameters $\mathbf{\theta}$ and training parameters
$\xi_{ij}$. This implies that the obtained optimization problem is quadratic
with linear constraints and has variables $\xi_{ij}$ and $\mathbf{\theta}$.

\section*{Numerical experiments}

In numerical experiments, the properties of the proposed SurvBESA model,
trained on synthetic and real data, are investigated. In synthetic
experiments, various models are compared, such as the single Beran model
(\textquotedblleft Single Beran\textquotedblright), the classical bagging of
Beran models with simple averaging of the predictions of weak models
(\textquotedblleft Bagging\textquotedblright), and the SurvBESA model, which
uses the self-attention mechanism, Huber $\epsilon$-contamination model, and
optimization. For real data, the models mentioned above are supplemented with
GBM Cox, GBM AFT, and RSF. The performance measure for studying and comparing
the models is the C-index, calculated on the test set. To evaluate the C-index
for each dataset, cross-validation is performed. Instances for training,
validation, and testing are randomly selected in each run. Hyperparameters are
tuned using the Optuna library.

\subsection{Synthetic data}

Training instances $\mathbf{x}\in \mathbb{R}^{5}$ are generated randomly based
on the uniform distribution within the hyperrectangle $\prod_{j=1}^{5}%
[a^{(j)},b^{(j)}]$. Censoring indicators $\delta$ are generated randomly
according to the Bernoulli distribution with probabilities $\Pr \{
\delta=0\}=1-p$ and $\Pr \{ \delta=1\}=p$. The number of instances $N$ in the
training set along with the Bernoulli distribution parameter are varied to
study their impact on the model performance. Event times are generated
according to the Weibull distribution with the shape parameter $k$. The event
times depend on $\mathbf{x}$ through the following relationship:%
\[
T=\frac{\sin \left(  c\cdot \sum_{i=1}^{5}x^{(i)}\right)  +c}{\Gamma
(1+1/k)}\cdot \left(  -\log(u)\right)  ^{1/k},
\]
where $c$ is a parameter, and $u$ is a random variable uniformly distributed
on the interval $[0,1]$, $x^{(i)}$ is the $i$-th feature of $\mathbf{x}$.

To make the synthetic data more complex, the expected value of $T$ is modified
according to the Weibull distribution as $\sin \left(  c\cdot \sum_{i=1}%
^{5}x^{(i)}\right)  +c$. In other words, the mean event time varies as a
sinusoidal function. The term $\sum_{i=1}^{5}x^{(i)}$ is used to model feature interactions.

Sets of the hyperparameter values for different models are as follows:

\begin{itemize}
\item \textbf{SurvBESA}: $\epsilon \in \lbrack0,1]$; $w\in \{10^{-3}%
,10^{-2},10^{-1},10^{0},10^{1},10^{2},10^{3}\}$; $\tau \in \{10^{-2}%
,10^{-1},10^{0},10^{1},10^{2},10^{3}\}$; the subset size varies in the range
$[0.1,0.7]$; the number of the Beran estimators is in the interval $[5,50]$;
the gradient descent step size is from the set$\{10^{-3},10^{-2},10^{-1}\}$.

\item \textbf{Bagging}: $\tau \in \{10^{-2},10^{-1},10^{0},10^{1},10^{2}%
,10^{3}\}$; the subset size varies in the range $[0.1,0.7]$; the number of the
Beran estimators is in $[5,50]$.

\item \textbf{Single Beran}: $\tau \in \{10^{-2},10^{-1},10^{0},10^{1}%
,10^{2},10^{3}\}$.
\end{itemize}

These hyperparameters, if not the subject of study, are determined using the
Optuna library \cite{akiba2019optuna}. Other hyperparameters are tested
manually, selecting those that yield the best results.

To investigate how the C-index depends on various parameters, corresponding
experiments are conducted. In all experiments for SurvBESA, the training
parameters are optimized using Adam with 100 epochs. The initial parameters of
the synthetic data are: $c=3$, $k=6$, $a^{(j)}=-2.0$, $b^{(j)}=5$,
$j=2,\dots,d$. The initial number of points $N$ in the training set is 200, in
the validation set is 100, and in the test set is 100, with the proportion of
censored data being 0.2. Some of the above parameters are varied in the
experiments to study their impact on the resulting C-index. For each value of
the investigated hyperparameters of the models and data generation parameters,
25 experiments were conducted, and their results were averaged.%

\begin{figure}
[ptb]
\begin{center}
\includegraphics[
height=2.8951in,
width=3.8538in
]%
{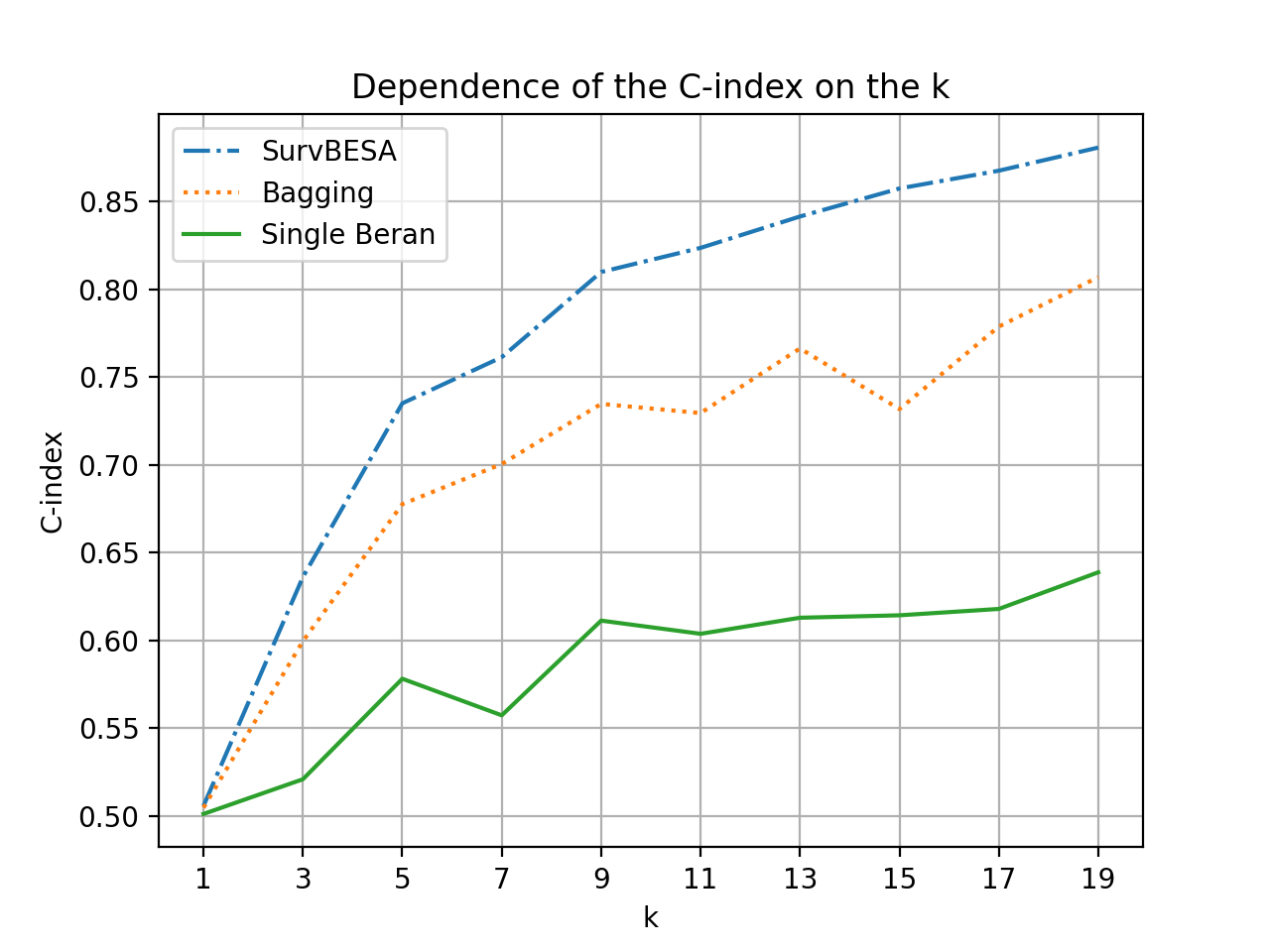}%
\caption{Dependence of the C-index on the parameter $k$}%
\label{f:k}%
\end{center}
\end{figure}

\emph{Dependence of the C-index on the parameter }$k$: The parameter $k$ is
varied in the set $[1,19]$ with a step size of 2. The study examined how the
C-index depends on the parameter $k$ of the Weibull distribution. In the case
where $k=1$, an exponential distribution is used for generation. If $k=2$, a
Rayleigh distribution is used. The larger the parameter $k$, the less noise in
the data. This can be observed in Fig. \ref{f:k}, which illustrates the
dependence of the C-index on the parameter $k$ for different models. The
C-index increases with increasing $k$, which is the expected behavior. At the
same time, the C-index for the SurvBESA model is higher for all values of $k$
compared to the Bagging and Single Beran models. Additionally, the C-index for
the Bagging model is higher than that of the single Beran model.

\begin{figure}
[ptb]
\begin{center}
\includegraphics[
height=2.8764in,
width=3.8294in
]%
{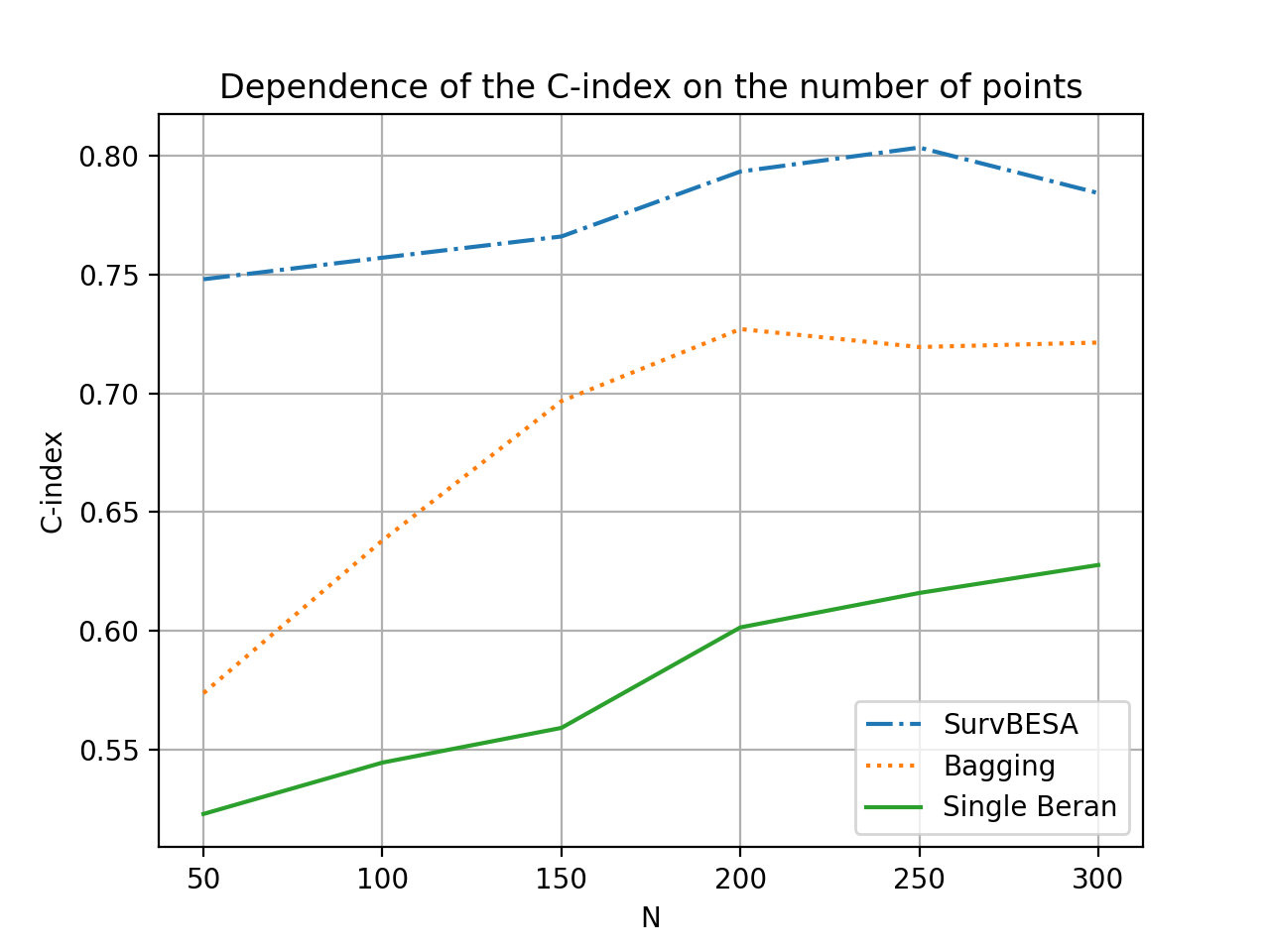}%
\caption{Dependence of the C-index on the number of points in the training
set}%
\label{f:course_points}%
\end{center}
\end{figure}

\emph{Dependence of the C-index on the number of points}: The next question is
how the number of points in the training set impact the model accuracy (the
C-index). The correpsonding results are depicted in Fig. \ref{f:course_points}.%

\begin{figure}
[ptb]
\begin{center}
\includegraphics[
height=2.7726in,
width=3.6893in
]%
{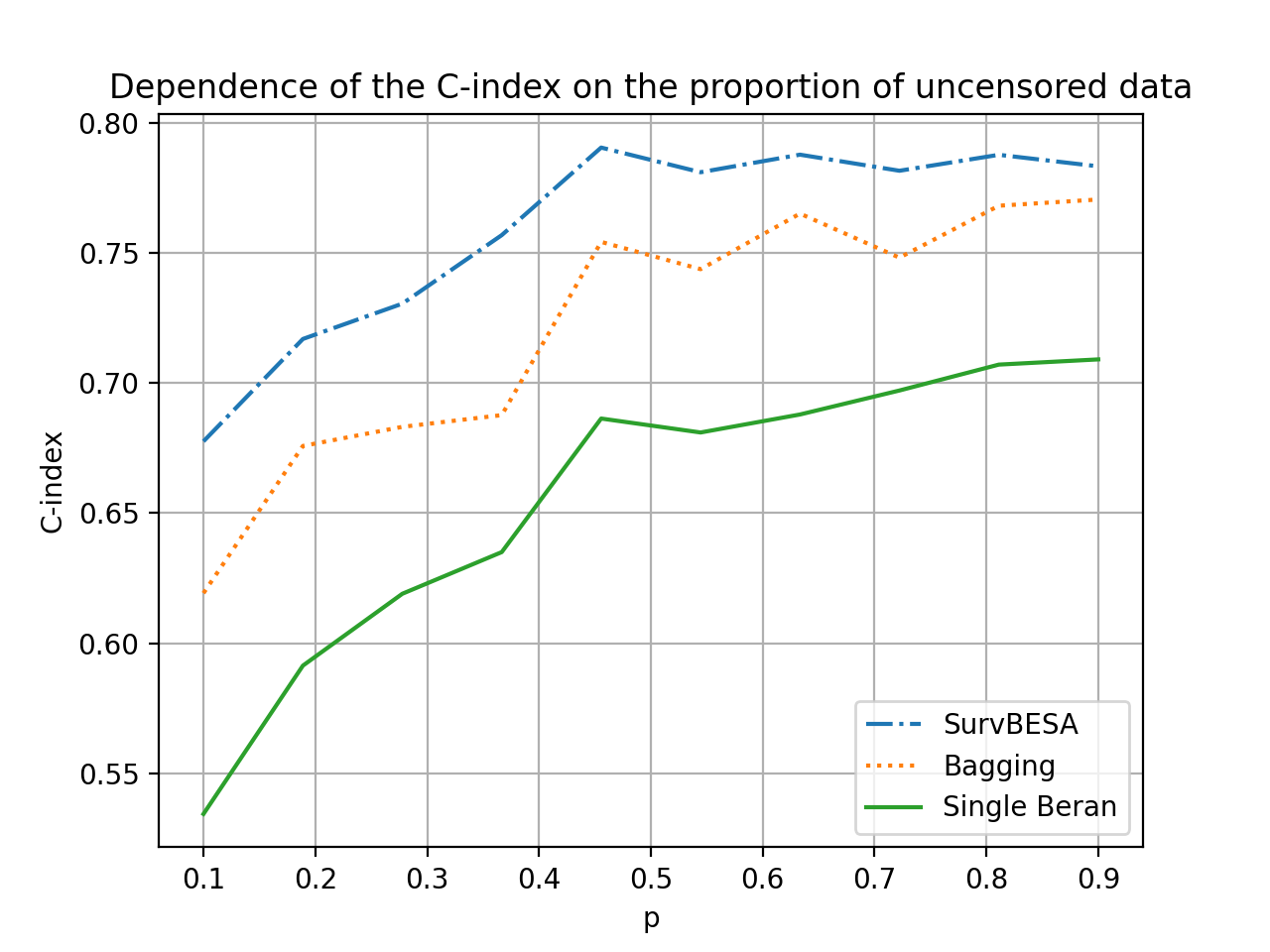}%
\caption{Dependence of the C-index on the proportion of uncensored data}%
\label{f:delta_obs}%
\end{center}
\end{figure}

\emph{Dependence of the C-index on the proportion of uncensored data}: Ten
values are chosen for the proportion of uncensored data $p$, starting from 0.1
and ending at 0.9, with equal steps. This experiment is necessary to study the
ability of models to learn in the presence of a large amount of censored data,
which is often encountered in real-world datasets. An interesting observation
is that as the proportion of uncensored data increases, the C-index values for
all models decrease. This can be observed in Fig. \ref{f:delta_obs}.%

\begin{figure}
[ptb]
\begin{center}
\includegraphics[
height=2.8399in,
width=3.7796in
]%
{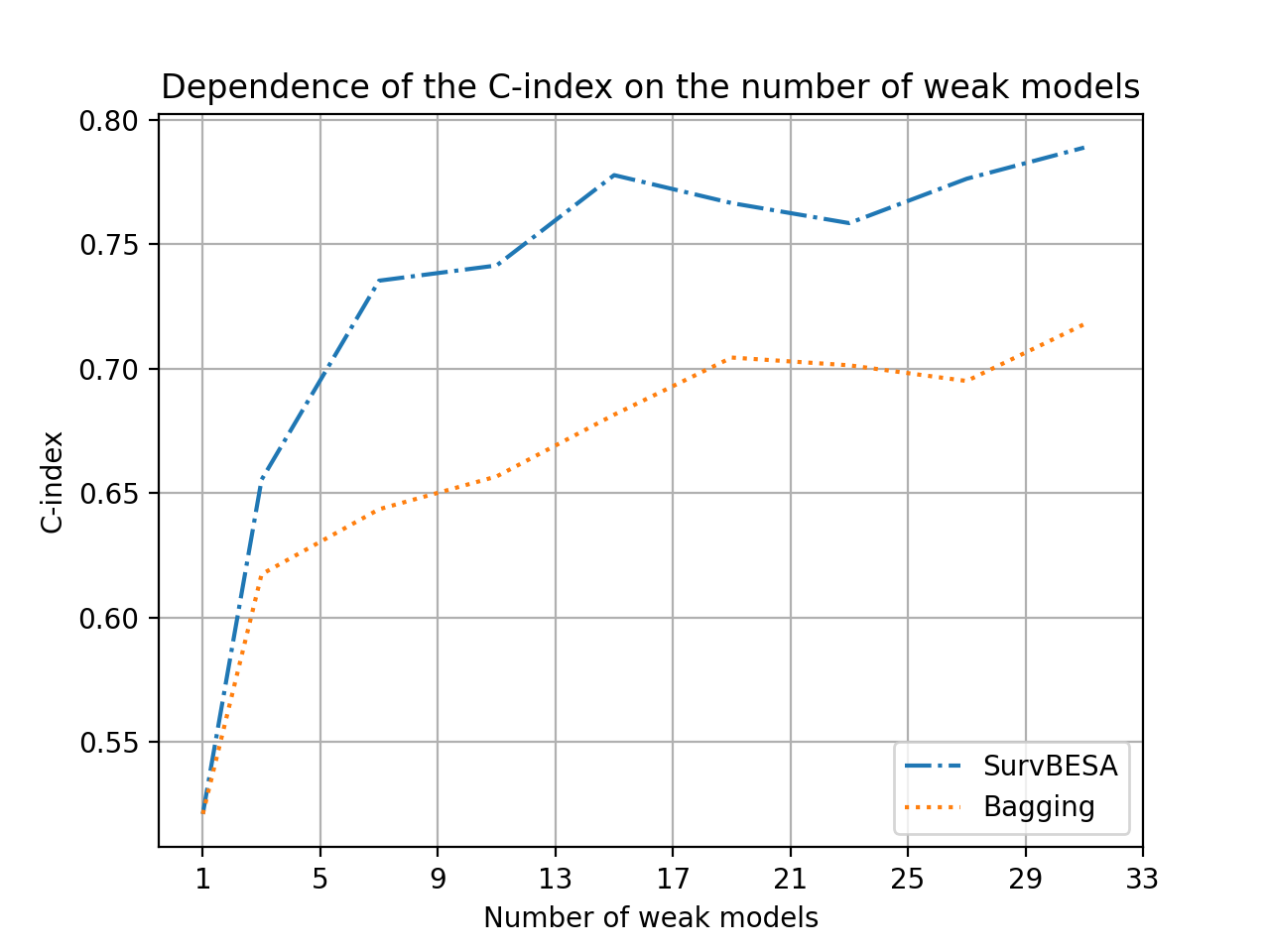}%
\caption{Dependence of the C-index on the number of weak models}%
\label{f:count_estimators}%
\end{center}
\end{figure}

\emph{Dependence of the C-index on the number of weak models}: The number of
weak models is varied in the set $[1,31]$ with a step size of 4, while the
subset size is fixed at 0.4 and not tuned. The case where the number of weak
models equals 1 is equivalent to the Single Beran model. The dependence is
shown in Fig. \ref{f:count_estimators}. It can be observed that for both
models, the C-index increases with the number of weak models, which aligns
with the core idea of bagging.%

\begin{figure}
[ptb]
\begin{center}
\includegraphics[
height=2.8893in,
width=3.8455in
]%
{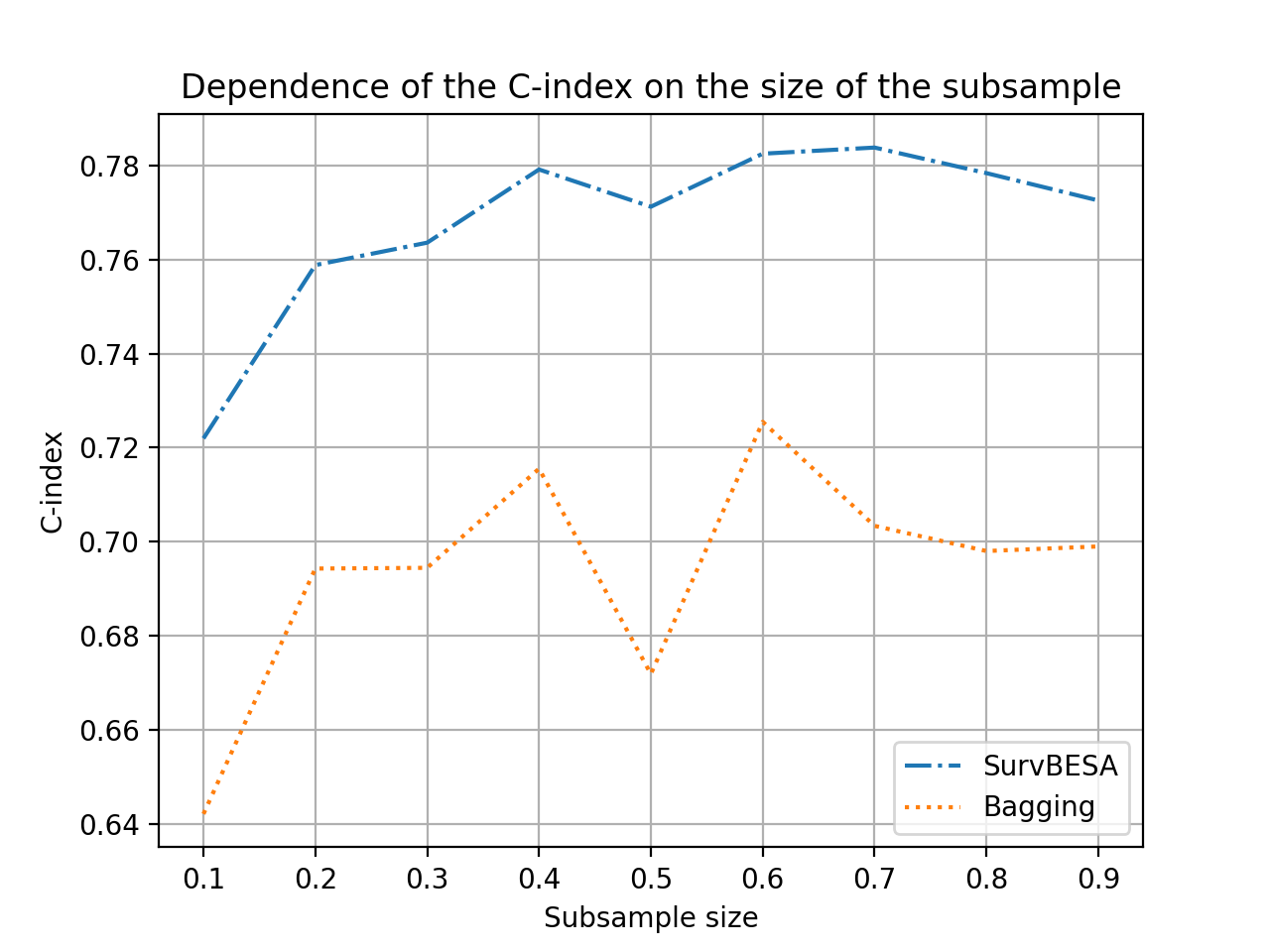}%
\caption{Dependence of the C-index on the subset size}%
\label{f:subsample_size}%
\end{center}
\end{figure}

\emph{Dependence of the C-index on the subset size}: The subset size is varied
in the range $[0.1,0.9]$. The number of weak models is fixed at 25. The
dependence is shown in Fig. \ref{f:subsample_size}. From the graph, it can be
seen that the C-index increases with the subset size and then plateaus. This
can be attributed to the fact that the parameter $\tau$ is tuned using the
training data that is not included in the Beran model. Additionally, the
random subspace partitioning used in building the ensemble may play a
significant role.%

\begin{figure}
[ptb]
\begin{center}
\includegraphics[
height=2.9025in,
width=3.8628in
]%
{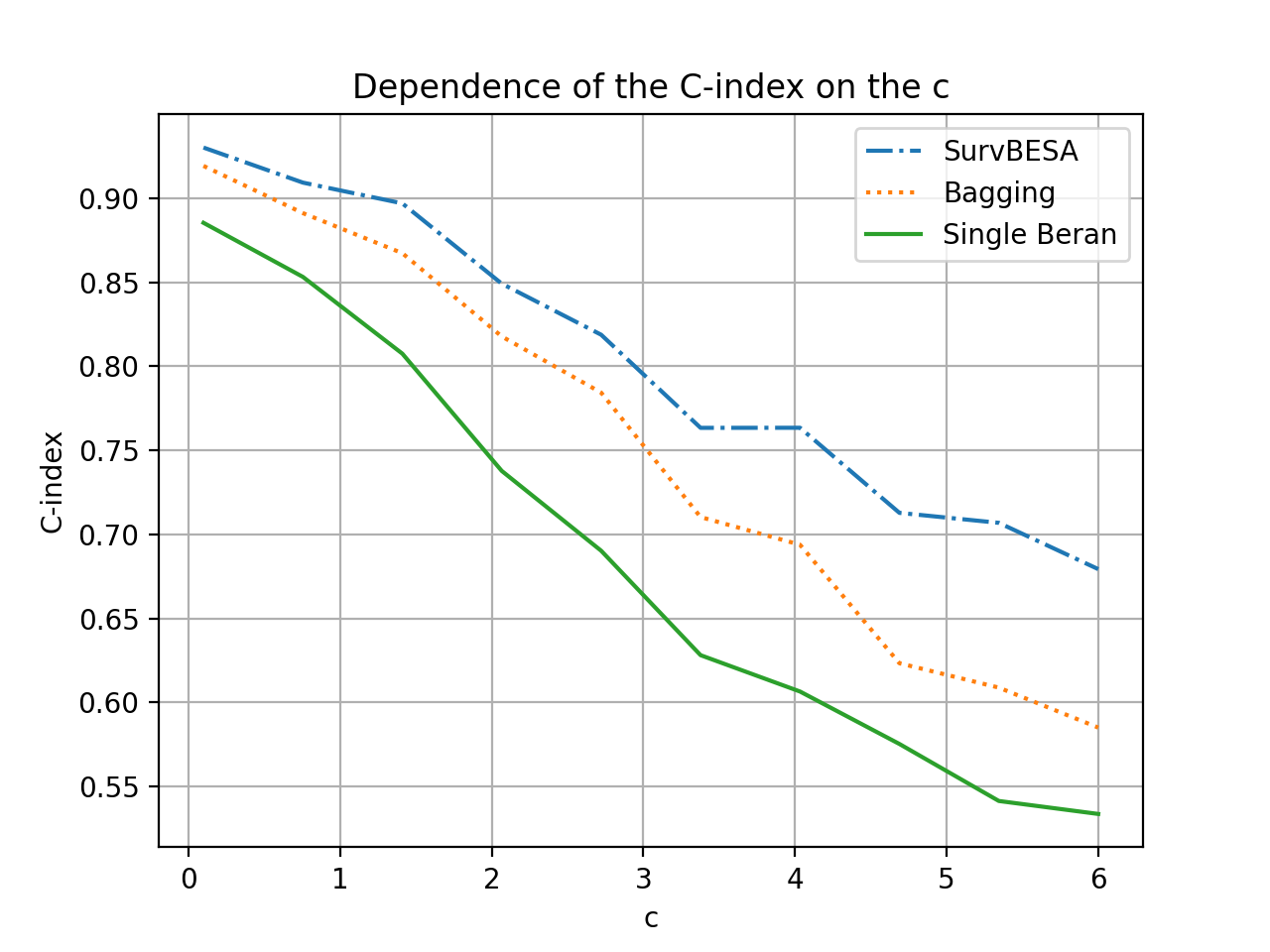}%
\caption{Dependence of the C-index on the parameter $c$}%
\label{f:c}%
\end{center}
\end{figure}

\emph{Dependence of the C-index on the parameter }$c$: Ten values are chosen
for the parameter $c$, starting from $0.1$ and ending at $6$, with equal
steps. The larger the parameter $c$, the \textquotedblleft
faster\textquotedblright \ the expected time changes, making its accurate
prediction more challenging. This is consistent with the results shown in Fig.
\ref{f:c}. The C-index decreases as the parameter $c$ increases. It is also
worth noting that the more complex the function relating the expected time to
the features $\mathbf{x}$, the greater the difference between SurvBESA and the
other models. In other words, as the complexity of the synthetic dataset
increases, the C-index for the SurvBESA model decreases more slowly compared
to the other models.%

\subsection{Real data}

To compare SurvBESA with various survival models on real data, the models
described above are used. They are tested on the following real benchmark
datasets: \emph{Veterans}: 137 examples, 6 features; \emph{AIDS}: 2139
examples, 23 features; \emph{Breast Cancer}: 198 examples, 80 features;
\emph{WHAS500}: 500 examples, 14 features; \emph{PBC}: 418 examples, 17
features; \emph{WPBC}: 198 examples, 34 features; \emph{HTD}: 69 examples, 9
features; \emph{CML}: 507 examples, 7 features; \emph{Rossi}: 432 examples, 62
features; \emph{Lung Cancer}: 228 examples, 9 features.

To evaluate the results, 100 iterations of different splits of the dataset
into training, validation, and test sets are performed. The training set
contains 60\% of the examples, the validation set contains 20\%, and the test
set contains 20\%. Hyperparameters are tuned on the validation set, and the
C-index is measured on the test set. The average C-index values across
different iterations are the final results used for comparison.

SurvBESA is implemented using Python software. The corresponding software is
available at: \url{https://github.com/NTAILab/SurvBESA}.

Hyperparameters for comparison of models:

\begin{itemize}
\item \emph{SurvBESA}: $\epsilon \in \lbrack0,1]$; $w\in \{10^{-3},10^{-2}%
,10^{-1},10^{0},10^{1},10^{2},10^{3}\}$; $\tau \in \{10^{-2},10^{-1}%
,10^{0},10^{1},10^{2},10^{3}\}$; the subset size varies in the range
$[0.1,0.7]$; the number of estimators is in $[5,50]$; the gradient descent
step size is in $\{10^{-3},10^{-2},10^{-1}\}$.

\item \emph{Bagging}: $\tau \in \{10^{-2},10^{-1},10^{0},10^{1},10^{2},10^{3}%
\}$; the subset size varies in the range $[0.1,0.7]$; the number of estimators
is in $[5,50]$.

\item \emph{Single Beran}: $\tau \in \{10^{-2},10^{-1},10^{0},10^{1}%
,10^{2},10^{3}\}$.

\item \emph{RSF}: the number of weak estimators varies in the range
$[20,300]$; the maximum tree depth varies in the range $[2,15]$; the minimum
number of samples in a leaf node varies in the range $[2,15]$.

\item \emph{GBM Cox}: the gradient descent step size varies in $\{10^{-3}%
,10^{-2},10^{-1}\}$; the number of iterations varies in the range $[20,300]$;
the maximum tree depth varies in the range $[2,10]$.

\item \emph{GBM AFT}: the gradient descent step size varies in $\{10^{-3}%
,10^{-2},10^{-1}\}$; the number of iterations varies in the range $[20,300]$;
the maximum tree depth varies in the range $[2,10]$.
\end{itemize}

The comparison results of different models for the above datasets are
presented in Table \ref{t:Table1} where the best results for each dataset are
highlighted in bold. It can be observed that SurvBESA outperforms other models
on most datasets. It is also important to note that the bagging Beran models
perform better than a single Beran model but does not always outperform RSF,
GBM Cox, and GBM AFT models. Although the results of SurvBESA are better than
those of other models, SurvBESA is significantly more computationally complex,
which may be important in some applications.%

\begin{table}[tbp] \centering
\caption{Results of comparison using the C-index metric for methods: ensemble of Beran models with simple averaging, single Beran model, SurvBESA, RSF, GBM Cox, and GBM AFT on various datasets}%
\begin{tabular}
[c]{|l|c|c|c|c|c|c|}\hline
Dataset & SurvBESA & Bagging & Single Beran & RSF & GBM Cox & GBM AFT\\ \hline
Veteran & $\mathbf{0.7414}$ & $0.6835$ & $0.6346$ & $0.7001$ & $0.6821$ &
$0.7004$\\ \hline
AIDS & $\mathbf{0.7686}$ & $0.7104$ & $0.6466$ & $0.7339$ & $0.7198$ &
$0.5800$\\ \hline
Breast Cancer & $\mathbf{0.7317}$ & $0.6706$ & $0.6595$ & $0.6806$ & $0.7021$
& $0.6979$\\ \hline
WHAS500 & $\mathbf{0.7730}$ & $0.7374$ & $0.7181$ & $0.7610$ & $0.7528$ &
$0.7492$\\ \hline
WPBC & $\mathbf{0.7652}$ & $0.6896$ & $0.6495$ & $0.6387$ & $0.6159$ &
$0.6078$\\ \hline
HTD & $\mathbf{0.8315}$ & $0.7864$ & $0.7421$ & $0.7809$ & $0.7749$ &
$0.7932$\\ \hline
Lung & $\mathbf{0.6875}$ & $0.6430$ & $0.6256$ & $0.6115$ & $0.5799$ &
$0.5588$\\ \hline
Rossi & $\mathbf{0.6243}$ & $0.5863$ & $0.5245$ & $0.5182$ & $0.5117$ &
$0.5380$\\ \hline
PBC & $0.8152$ & $0.7712$ & $0.7619$ & $0.8099$ & $\mathbf{0.8153}$ &
$0.7969$\\ \hline
CML & $\mathbf{0.7149}$ & $0.7050$ & $0.7009$ & $0.7129$ & $0.7136$ &
$0.7114$\\ \hline
\end{tabular}
\label{t:Table1}%
\end{table}%

To formally demonstrate that SurvBESA surpasses other methods, we employ the
$t$-test, as proposed by Demsar \cite{Demsar-2006}, to assess whether the mean
difference in performance between two classifiers is significantly different
from zero. In this case, the $t$ statistic follows the Student's
$t$-distribution with $10-1$ degrees of freedom. The computed $t$ statistics
yield the corresponding $p$-value. The first sample consists of the C-indices
obtained from SurvBESA (see Table \ref{t:Table1}) whereas the second sample
consists of the best results provided by other models. The test show that
SurvBESA significantly outperforms other analyzed models, as the corresponding
$p$-value is equal to $0.0017$, i.e., it is smaller than $0.05$.

\subsection{Additional experiments with real data}

\begin{figure}
[ptb]
\begin{center}
\includegraphics[
height=2.4377in,
width=3.2439in
]%
{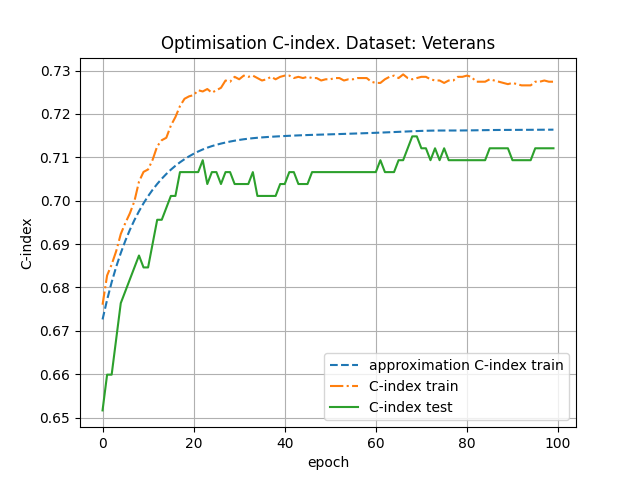}%
\caption{Illustration of the SurvBESA training process for the Veteran
dataset}%
\label{f:optimisation_veterans}%
\end{center}
\end{figure}

\begin{figure}
[ptb]
\begin{center}
\includegraphics[
height=2.48in,
width=3.3016in
]%
{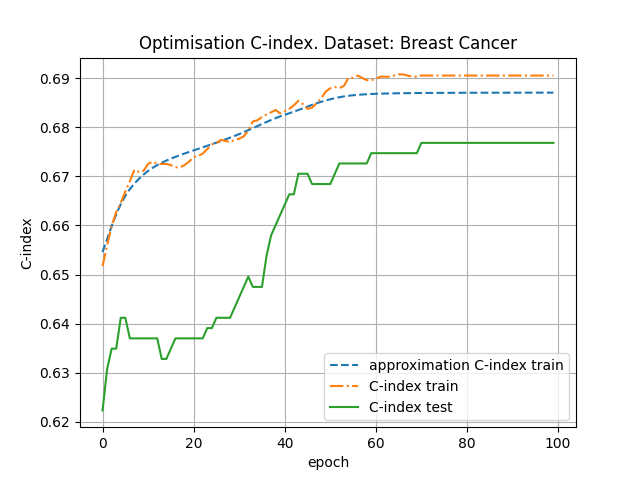}%
\caption{Illustration of the SurvBESA training process for the Breast Cancer
dataset}%
\label{f:optimisation_bc}%
\end{center}
\end{figure}

\begin{figure}
[ptb]
\begin{center}
\includegraphics[
height=2.498in,
width=3.3259in
]%
{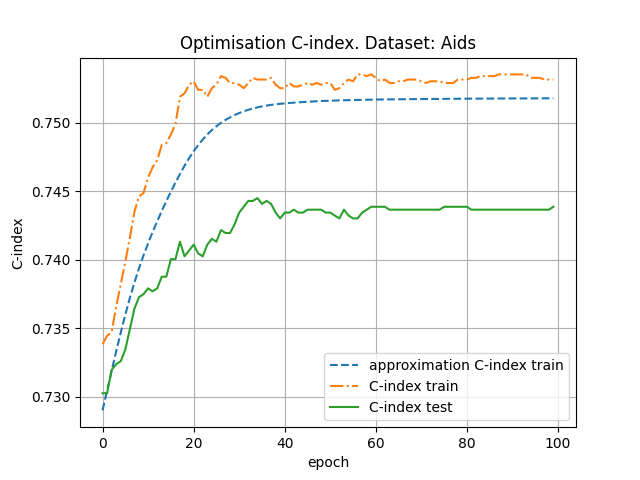}%
\caption{Illustration of the SurvBESA training process for the AIDS dataset}%
\label{f:optimisation_aids}%
\end{center}
\end{figure}

Figs. \ref{f:optimisation_veterans}-\ref{f:optimisation_aids} demonstrate the
gradient descent training process of the SurvBESA model. The datasets used are
Veteran, AIDS, and Breast Cancer. The graphs show the change in the C-index
with each epoch for the training and test sets, as well as the approximation
of the C-index using the sigmoid function proposed in this work. The C-index
approximation serves as the loss function, which is optimized using the Adam
method with the learning rate $0.1$. It is evident that with each iteration,
the C-index for both the training and test sets increases and eventually
plateaus, indicating the model's ability to learn.%

%

%

\begin{figure}
[ptb]
\begin{center}
\includegraphics[
height=3.9246in,
width=3.2122in
]%
{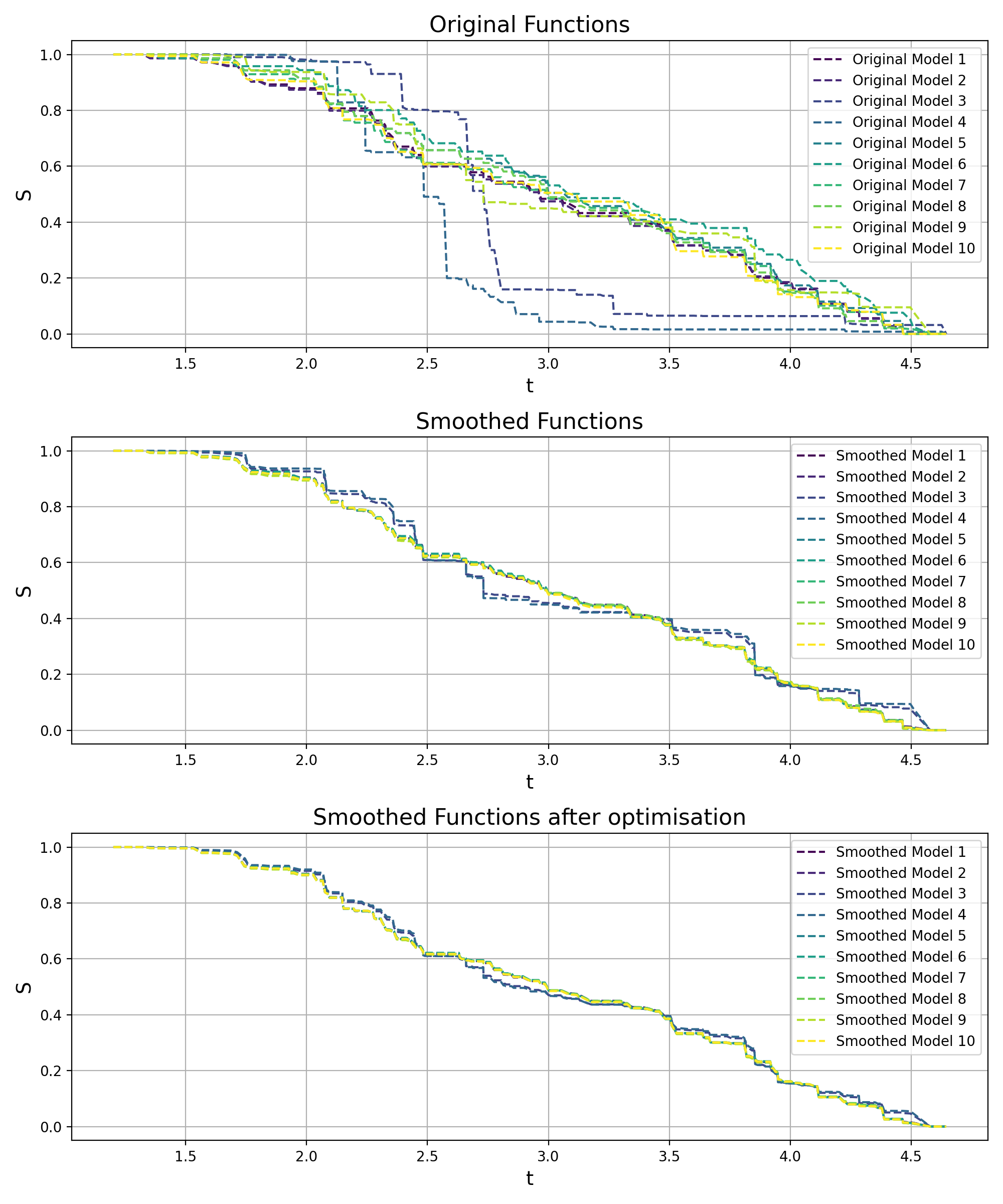}%
\caption{An illustrative example of the SF transformation produced by the weak
Beran models using the self-attention mechanism}%
\label{f:sfs}%
\end{center}
\end{figure}

Fig. \ref{f:sfs} shows the change in the SFs obtained from the Beran models as
a result of using the self-attention mechanism and optimization. It can be
observed that initially, some Beran models (models 3 and 4 in the first
picture in Fig. \ref{f:sfs}) produce SFs that significantly differ from the
others and can be viewed as outliers, leading to a certain bias in the final
predicted result. After applying self-attention (the second picture in Fig.
\ref{f:sfs}) and optimization (the third picture in Fig. \ref{f:sfs}), the SFs
are transformed into a more general form.%

\section{Conclusion}

In this paper, a novel ensemble-based survival model called SurvBESA has been
introduced. It leverages the Beran estimator as its weak learner and
incorporates the self-attention mechanism to aggregate predictions from
multiple base models. The proposed model addresses key challenges in survival
analysis, such as handling censored data and improving the robustness of
predictions in the presence of multi-cluster data structures. By applying
self-attention to the survival functions predicted by individual Beran
estimators, SurvBESA effectively reduces noise and stabilizes predictions,
leading to more accurate and reliable survival estimates.

Extensive numerical experiments on both synthetic and real-world datasets
demonstrate the superiority of SurvBESA over traditional survival models,
including Random Survival Forests (RSF), Gradient Boosting Machines (GBM) with
Cox and AFT loss functions, and standalone Beran estimators. The results
highlight SurvBESA's ability to adapt to complex data structures and its
robustness in scenarios with high levels of censoring. Furthermore, the
model's performance is consistently strong across various datasets, making it
a versatile tool for survival analysis tasks.

A key contribution of this work is the integration of the Huber $\epsilon
$-contamination model into the self-attention framework, which simplifies the
training process by reducing it to a quadratic or linear optimization problem.
This use of the $\epsilon$-contamination model not only enhances computational
efficiency, but also provides a principled way to handle uncertainty in the
predictions of weak learners.

The implementation of SurvBESA is publicly available, enabling researchers and
practitioners to apply and extend the model to their own survival analysis
problems. Future work could explore further optimizations of the
self-attention mechanism, extensions to other types of weak learners, and
applications to larger and more diverse datasets.

In conclusion, SurvBESA represents a significant advancement in ensemble-based
survival analysis, offering a robust, accurate, and generalizable framework
for predicting time-to-event outcomes in the presence of censored data. Its
ability to leverage the strengths of the Beran estimator while mitigating its
limitations through self-attention makes it a promising tool for both research
and practical applications in survival analysis.

While the Beran estimator has proven effective as a weak learner,
investigating the integration of other types of weak learners into the
SurvBESA framework could broaden its applicability. For instance,
incorporating neural network-based survival models or other non-parametric
estimators could provide additional flexibility and performance gains.

Future research could focus on refining the self-attention mechanism within
SurvBESA to further enhance its ability to capture dependencies among survival
function predictions. This could involve exploring alternative attention
architectures, such as multi-head attention or sparse attention, to improve
computational efficiency and prediction accuracy.

\section{Acknowledgement}

The research is partially funded by the Ministry of Science and Higher
Education of the Russian Federation as part of state assignments
\textquotedblleft Development and research of machine learning models for
solving fundamental problems of artificial intelligence for the fuel and
energy complex\textquotedblright \ (topic FSEG-2024-0027).

\bibliographystyle{unsrt}
\bibliography{Attention,Boosting,Classif_bib,Deep_Forest,Explain,Explain_med,MYBIB,MYUSE,Survival_analysis}

\end{document}